\documentclass{article}

\usepackage{arxiv}

\usepackage[utf8]{inputenc} 
\usepackage[T1]{fontenc}    
\usepackage{hyperref}       
\usepackage{url}            
\usepackage{booktabs}       
\usepackage{amsfonts}       
\usepackage{nicefrac}       
\usepackage{microtype}      
\usepackage{lipsum}
\usepackage{graphicx}
\usepackage{amsmath}  
\usepackage{amssymb}
\usepackage{multirow}   
\usepackage{tabularx}
\usepackage{algorithmicx, algpseudocode} 
\usepackage{algorithm}         
\usepackage{subcaption}
\usepackage{siunitx}
\usepackage{float}
\graphicspath{ {./images/} }

\title{CLIP-RLDrive: Human-Aligned Autonomous Driving via CLIP-Based Reward Shaping in Reinforcement Learning}

\author{
 Erfan Doroudian \\
  Department of Mechanical, Industrial and Aerospace Engineering\\
  Concordia University\\
  Montreal, QC \\
  \texttt{erfan.doroudian@concordia.ca} \\
   \And
 Hamid Taghavifar \\
  Department of Mechanical, Industrial and Aerospace
  Engineering\\
  Concordia University\\
  Montreal, QC \\
  \texttt{hamid.taghavifar@concordia.ca} \\
  }

\begin{document}
\maketitle
\begin{abstract}

This paper presents CLIP-RLDrive, a new reinforcement learning (RL)-based framework for improving the decision-making of autonomous vehicles (AVs) in complex urban driving scenarios, particularly in unsignalized intersections. To achieve this goal, the decisions for AVs are aligned with human-like preferences through Contrastive
Language-Image Pretraining (CLIP)-based reward shaping. One of the primary difficulties
in RL scheme is designing a suitable reward model, which can often be challenging to
achieve manually due to the complexity of the interactions and the driving scenarios. To deal with this issue, this paper leverages Vision-Language Models (VLMs), particularly CLIP, to build an additional reward model based on visual and textual cues.
CLIP’s ability to align image and text embeddings provides important features for translating human-like instructions into reward signals to guide the AV’s decision-making process. In addition, two RL algorithms are applied, Proximal Policy Optimization (PPO) and Deep Q-Network (DQN), to train an agent in complex unsignalized intersection environments. The performance of these algorithms is compared
with and without the CLIP-based reward model, which emphasizes the effect of CLIP on the agent’s ability to learn and optimize its behavior in a way that aligns with desired driving actions. Our obtained results demonstrate that the CLIP-based DQN achieves a 96\% success rate with only a 4\% collision rate, significantly outperforming CLIP-based PPO, which achieves a success rate of 38\% with a 54\% timeout rate. This is suggestive of the superior effectiveness of the proposed framework in aligning AV behavior with human-like driving standards.
\end{abstract}


\section{\textsc{Introduction}}

\subsection{\textsc{Background}}
The successful deployment of autonomous vehicles (AVs) into road systems, particularly in urban settings, demands both effective navigation and control capabilities, as well as the ability to emulate human driving behaviors \cite{huang2021driving}. To achieve this level of human-like performance, social cognition is needed in terms of understanding complex social interactions while navigating and interpreting human-like decision-making and social dynamics. This human-centric approach extends beyond environmental perception and serves as a much-needed step forward to make interaction with human drivers, pedestrians, and other road users safe and reliable \cite{taghavifar}. One of the main obstacles regarding the large-scale commercialization of fully autonomous systems from industries such as Waymo and Tesla is the handling of edge cases \cite{cui2023drivellm}, which refer to rare or unconventional situations that fall outside the scope of the scenarios typically encountered during training or testing. These scenarios are particularly difficult because they include complex interactions or unexpected conditions that deviate from standard traffic rules or patterns, such as adverse weather conditions, irregular behaviors of road users, and ambiguous traffic situations. As a result, traditional decision-making systems struggle to generalize or adapt effectively due to their rigid design. In contrast, a human driver outperforms in these edge cases due to the unique human ability to use intuition, social intelligence, and reasoning. Hence, it is critical to be able to connect the rule-based decision-making systems and the intuitive interpretation ability inherent in human driving behaviors \cite{taghavifar2024socially}.

Although recent progress observed in Reinforcement Learning (RL), deep learning, and neural networks (NNs) have contributed to improved decision-making abilities of AVs in complex and crowded urban environments \cite{taghavifar2024socially, toghi2022social, tram2019learning, ronecker2019deep}, they still struggle with long-tail scenarios due to the limited scope of patterns within their training datasets \cite{li2024large}. On the other hand, the recent developments in Large Language Models (LLMs) have presented a transformational opportunity for improving RL frameworks. LLMs can be used to generate context-aware instructions or explanations that provide AVs with additional guidance in complex traffic scenarios \cite{llms}. These models can interpret human preferences, predict consequences, and even explain the rationale behind decisions, and therefore build a hybrid method that improves the raw data-driven learning through human-like reasoning. Hence, the decision-making process of AVs can be enhanced to include both the optimal performance and also socially and ethically aligned actions that are interpretable to other road users as well \cite{cui2023drivellm, li2024large, li2023towards, zheng2024planagent}.

In RL, the agent's objective is to maximize cumulative returns, which makes the design of the reward function a critical and often challenging task \cite{ardashir}. Because the optimal policy is inherently defined by this reward function, sparse or delayed rewards in real scenarios can significantly reduce the learning of the agents \cite{kiran2021deep}. In addition, because RL agents depend only on reward signals to optimize their actions, slow or limited feedback can also prevent good learning progress as well. To remedy this drawback, additional guidance in the form of a shaping reward can be introduced, which supplements the environment’s natural reward signal and thus improves learning speed and performance \cite{kwon2023reward, chan2024dense}. This process originates from experimental psychology, where it involves reinforcing all actions that contribute toward the desired behavior \cite{kiran2021deep}. While reward shaping can accelerate learning by providing additional guidance, it may also cause the agent to optimize an augmented reward function $R^{\prime}$ instead of the original reward $R$. Hence, the resulting policy might be suboptimal for the intended objective, which is because the agent might end up with behaviors that maximize the shaping reward in ways that are different from, or even in conflict with, the original task goal defined by $R$. For example, in Ref. \cite{randlov1998learning}, the experimented bicycle agent returned in a circle to stay upright rather than reach its goal, which illustrated how reward shaping can lead agents to prioritize the shaping reward over the original objective. 

Several techniques have been proposed to address the potential issues with reward shaping. These methods are specifically designed to provide the agent’s behavior alignment with the original task objectives, which also prevents it from becoming disproportionately influenced by the additional shaping rewards. Recently, leveraging pre-trained foundation models to generate reward signals for RL adjustment has become a needed approach in the development of LLMs \cite{bai2022constitutional}. Some approaches only require a small amount of natural language feedback instead of a whole dataset of human preferences \cite{scheurer2023training}.  Although the use of language models has become popular to compute a reward function from a structured environment representation \cite{xie2023text2reward}, many RL tasks are indeed visual and demand using Vision-Language Models (VLMs) instead.

\subsection{\textsc{Related Works}}
LLMs have demonstrated promising performance across a diverse range of natural language tasks \cite{devlin2018bert, achiam2023gpt, touvron2023llama}. They can be an effective method for reward shaping by utilizing human feedback in the system's learning process, where this feedback helps define the task and guide the agent's behavior \cite{NIPS2017_d5e2c0ad}. Reinforcement Learning from Human Feedback (RLHF) is a technique where human feedback is used to guide the training process of an RL agent, especially when designing a reward function is difficult. Bai \textit{et al.} \cite{bai2022constitutional} explored the use of RLHF to calibrate LLMs and showed its ability to align AI behavior with human preferences for helpfulness and harmlessness while improving NLP performance through iterative feedback and preference modeling. However, collecting high-quality human feedback can be expensive and laborious. 

Therefore, the use of pre-trained models to determine the reward signal has shown as an effective strategy in research on LLMs. LLMs' human-like reasoning features are suggestive of their suitability in agent decision-making \cite{wang2024rl}. For example, Yu \textit{et al.} \cite{yu2023language} introduced a novel paradigm that utilized LLMs to generate reward functions from high-level language instructions, which as a result, enabled intuitive and flexible interaction with robots. Additionally, Cui \textit{et al.} \cite{cui2023drivellm} employed a hybrid decision-making framework combining LLM-based and rule-based systems to balance adaptability and safety. They demonstrated the ability of their framework to manage edge cases, such as snowy intersections and animal crossings, with superior contextual awareness compared to traditional methods. The GPT-Driver \cite{mao2023gpt} framework also introduced a novel perspective on motion planning for AVs by utilizing LLMs such as GPT-3.5 to address the limitations of traditional methods. Their proposed method reformulated motion planning as a language modeling task by converting heterogeneous inputs such as perception data and ego-states into language tokens. Such addition can allow the model to generate waypoint-based trajectories through natural language descriptions \cite{sadigh2}.
However, designing a suitable reward function is still complex for various types of tasks. 

Many RL tasks are visual and need to use Vision-Language Models (VLMs) instead. Rocamonde \textit{et al.} \cite{rocamonde2023vision} introduced a method using CLIP as a zero-shot reward model for reinforcement learning, which helps agents learn complex tasks such as humanoid movements from simple text prompts. This approach removes the demand for manual reward design. In Ref. \cite{adeniji2023language} Language Reward Modulated Pretraining (LAMP) on RLBench was evaluated using a robotic manipulation benchmark setup, which showed significant improvements in sample efficiency and the ability to deliver tasks such as \textit{Pick Up Cup} and \textit{Push Button}. In addition, several key challenges in traditional RL methods were introduced in Ref. \cite{adeniji2023language}, such as dependency on handcrafted rewards and challenges in adapting to diverse downstream tasks.
In Ref. \cite{wang2024rl}, by assigning VLMs to compare agent observations based on task descriptions, their proposed method eliminated the need for human-labeled rewards or access to low-level ground-truth states.

The reviewed literature indicates that while LLMs have recently been suggestive of an enhanced decision-making process, the combination of VLMs and RL is still largely unexplored, especially considering the visual interactions of agents within their environment. In particular, our paper, compared to the existing literature, proposes a pre-trained CLIP on a custom dataset using transfer learning techniques to improve the model's performance in the specified intersection scenario. In addition, the present study aims to develop and evaluate a novel framework that combines RL with VLMs to improve AV decision-making in unsignalized intersections, with a focus on mimicking human-like behaviors. The challenge of managing AV behavior at intersections without traffic signals requires an understanding of contextual, human-like decision-making processes. In addition, this study aims to enable the AV to make safe and efficient decisions in response to instantaneous visual inputs by using CLIP \cite{radford2021learning} as a VLM to build a reward model. Therefore, the three main contributions in this paper can be summarized in the following:

 \begin{itemize}
     \item A CLIP-based reward model is introduced for guiding AV behavior. By optimizing it, this research uses CLIP’s contextual understanding to provide instantaneous guidance to the AV to align its actions with safe and human-inspired behaviors.
     \item A set of data (visual scenario, description) pair is collected to modify the CLIP using our dataset. To optimize the training process, this study utilizes transfer learning by optimizing only the upper layer of CLIP while keeping the remaining layers unchanged. This approach reduces computational demands, which provides efficient training with a limited dataset and enhances the practicality of the framework.

     \item This research comprises CLIP's recommendations as a secondary reward signal, combined with the environment’s basic reward function, within the RL framework. This dual reward structure guarantees that the AV's actions both maximize traditional rewards and align with CLIP's human-like guidance to improve decision-making quality.
 
 \end{itemize}

Finally, the effectiveness of the proposed framework is demonstrated using the \textit{Highway-env} simulation environment \cite{highway-env}, specifically its intersection scenario, which is implemented in OpenAI Gym \cite{brockman2016openai}. 
 
The organization of the remainder of this paper is as follows: In Section \ref{s2}, the training of an RL agent at an unsignalized intersection as a partially observable Markov decision process (POMDP) is defined. In addition, the observation and action spaces are outlined, and CLIP as a reward model is included to guide human-like decision-making using DQN and PPO algorithms.  Section \ref{s3} presents the solution setup and the main results achieved so far. Finally, Section \ref{s4} concludes the paper and suggests directions for future research.

\section{\textsc{Problem Formulation}}
\label{s2}
The problem of training RL agent in an unsignalized intersection is formulated as a POMDP by the tuple $\mathcal{M}:= \langle \mathcal{S}, \mathcal{A}, \Omega, \mathcal{T}, \mathcal{O}, \mathcal{R}, \gamma \rangle$, which is a form of fully observable Markov Decision Process (MDP) that the agent perceives an observation $o \in \Omega$,  while it lacks complete access to the actual state of the environment. $\mathcal{S}$ is a finite set of states, \(\mathcal{A}\) is a finite set of actions,  \(\Omega\) is a finite set of observations, \(\mathcal{T}\) is a transition probability defined as  \(\mathcal{T}: \mathcal{S} \times \mathcal{A} \times \mathcal{S} \to [0,1]\), 
\(\mathcal{O}\) is an observation function defined as  \(\mathcal{O} : \mathcal{S} \times \mathcal{A} \times \Omega \to [0,1]\),  \(\mathcal{R}\) is a reward function defined as 
\(\mathcal{R} : \mathcal{S} \times \mathcal{A} \times \mathcal{S} \to \mathbb{R}\), and $\gamma$ is the discount factor. A sequence
of states and actions is called a trajectory $\tau = \langle s_0, a_0, s_1, a_1, \dots \rangle$, where $s_i \in \mathcal{S}$, $a_i \in \mathcal{A}$, and the length of the trajectory $|\tau|$ is considered finite. Thus, the primary objective is to identify a policy $\pi(s, a)$ that maximizes the discounted sum of future rewards over an infinite time horizon $\pi^*:\mathcal{S} \to \mathcal{A}$. 
\begin{equation}
    \pi^* := \arg \max_{\pi} \mathbb{E}_{\tau \sim \pi} \left[ \sum_{t=0}^{T} \gamma^t r_t(s_t, \pi(s_t))\right]
\end{equation}
where $\mathbb{E}$ denotes the expectation operator, and $\gamma \in [0,1)$ is the discount factor. To find the optimal policy $\pi ^ *$, Deep $Q$-Network (DQN) \cite{mnih2013playing}, and Proximal Policy Optimization (PPO) \cite{schulman2017proximal} are utilized in this paper as a value-based, and policy-based deep reinforcement learning (DRL) method, respectively.

\subsection{\textsc{Deep $Q$-Network}}
In RL, the state-action value function \( Q^\pi(s, a) \) under a policy \( \pi \) represents the expected cumulative reward when starting from state \( s \), taking action \( a \), and following policy \( \pi \) thereafter, formulated as follows:
\begin{equation}
   Q^\pi(s, a) = \mathbb{E}_\pi \left[ \sum_{t=0}^\infty \gamma^t r_t(s_t, \pi(s_t)) \mid s_0 = s, a_0 = a \right]
\end{equation}
The optimal policy maximizes the state-action value function, i.e., $\pi^*(s) = \arg \max_a Q^*(s, a)$, and the optimal state-action value function \( Q^*(s, a) \) that satisfies the Bellman equation, can then be derived as:
\begin{equation}
    Q^*(s, a) = \mathbb{E} \left[ r + \gamma \max_{a'} Q^*(s', a') \mid s_0 = s, a_0 = a \right]
\end{equation}
To approximate \( Q^*(s, a) \), DQN uses a neural network \( Q(s, a; \theta) \) parameterized by \( \theta \). Then it minimizes the mean squared Bellman error between the current $Q$-values and the target $Q$-values:
\begin{equation}
    \mathcal{L}(\theta) = \mathbb{E} \left[ \left( y - Q(s, a; \theta) \right)^2 \right]
\end{equation}
where the target \( y \) is defined as:
\begin{equation}
    y = r + \gamma \max_{a'} Q_{\text{target}}(s', a'; \theta^-)
\end{equation}
and the target network $Q_{\text{target}}(s', a'; \theta^-)$, is introduced to stabilize training by providing fixed $Q$-values as a reference during updates, where $\theta^-$ represents the target network parameters, which the weights from the main network $\theta$ are copied to the target network periodically.  Also, the network parameters $\theta$ are updated using stochastic gradient descent (SGD):
\begin{equation}
    \theta_{i+1} \leftarrow \theta_i - \alpha \nabla_\theta L(\theta_i)
\end{equation}

where $\nabla_\theta L(\theta)$ is the gradient of the loss function with respect to the network  $\theta$. Furthermore, DQN uses an experience replay mechanism to break the correlation between the training samples. Actions and states are stored in a replay buffer as tuples $(s_t, a_t, r_t, s_{t+1})$, and mini-batches of these tuples are randomly sampled to update the network. Also, exploration is managed by an $\epsilon$-greedy policy, where $\epsilon$ is gradually decreased over time.

\subsection{\textsc{Proximal Policy Optimization}}
\begin{figure}
    \centering
    \includegraphics[width=0.8\linewidth]{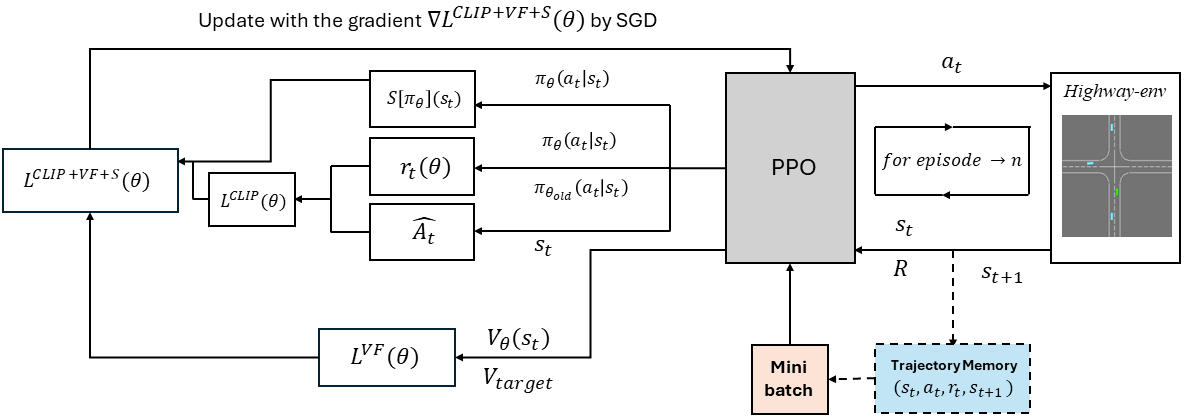}
    \caption{A general block diagram of the proposed framework based on PPO.}
    \label{fig:ppo}
\end{figure}

Unlike DQN, PPO is a policy-based method that directly optimizes the policy by maximizing the expected cumulative rewards. The policy \( \pi_\theta(a|s) \) in PPO is parameterized by a neural network with parameters \( \theta \). This network outputs a probability distribution over actions for each state \( s \). For discrete action spaces, the output is a softmax distribution:
\begin{equation}
    \pi_\theta(a|s) = \text{softmax}(\mathcal{Z}_a)
\end{equation}
where $\mathcal{Z}_a$ is the logit corresponding to action $a$. PPO optimizes a surrogate objective function to improve training stability while preventing large policy updates. This approach introduces a clipping mechanism that ensures the policy does not change too drastically during each training step. The objective function is shown as follows:
\begin{equation}
\mathcal{L}(\theta) = \mathbb{E}_t \left[ \min \left( r_t(\theta) \hat{A}_t, \text{clip}(r_t(\theta), 1-\epsilon, 1+\epsilon)  \hat{A}_t \right) \right]
\end{equation}
where $r_t(\theta) = \frac{\pi_\theta(a_t | s_t)}{\pi_{\theta_{\text{old}}}(a_t | s_t)}$ is the probability ratio, comparing the new policy $\pi_{\theta}$ to the old policy $\pi_{\theta_{\text{old}}}$, \(\epsilon\) is a small clipping parameter that constrains the probability ratio $r_t(\theta)$ within the range $[1-\epsilon, 1+\epsilon]$, and $\hat{A}_t$ is an estimator of the advantage function at time $t$. The advantage function is defined as follows:
\begin{equation}
    \hat{A}_t = \delta_t + (\gamma \lambda) \delta_{t+1} +\dots + (\gamma \lambda)^{T-t+1} \delta_{T-1}
\end{equation}
\vspace{0.1cm}

where $\delta_t = r_t + \gamma V(s_{t+1}) - V(s_t), r_t$ is the reward at time $t, V(s)$ is the value function at state $s$, $\gamma$ is a discount factor and $\lambda$ helps to balance the bias-variance trade-off in advantage estimation. The objective function encourages the policy to improve (when $A_t >0$) while preventing large updates that might degrade performance. This balance makes PPO simpler and more effective compared to methods, \textit{e.g.}, Trust Region Policy Optimization (TRPO) \cite{schulman2015trust}. In addition to optimizing the policy, PPO trains a value function \( V(s) \) to approximate the expected return from state \( s \). The value function is updated by minimizing the following loss:
\begin{equation}
        \phi_{k+1} = \arg \min_{\phi} \frac{1}{|\mathcal{D}_k| T} \sum_{\tau \in \mathcal{D}_k} \sum_{t=0}^{T} \left( V_{\phi}(s_t) - \hat{R}_t \right)^2
\end{equation}
where $\phi$ represents the parameters of the value function $V_{\phi}$, which is a neural network, $\mathcal{D}_k$ is a dataset of trajectories collected during iteration $k$, and $\hat{R}_t$ is reward-to-go at time $t$ computed as:
\begin{equation}
    \hat{R}_t = r_t + \gamma r_{t+1} + \gamma^2 r_{t+2} + \dots + \gamma^{T-t} r_T
\end{equation}
which is the actual observed cumulative reward from state $s_t$ onward. Fig. \ref{fig:ppo} illustrates a brief representation of the proposed framework. 

\subsection{\textsc{Action and Observation Spaces}}
\begin{figure}[!ht]
    \centering
    \includegraphics[width=0.75\linewidth]{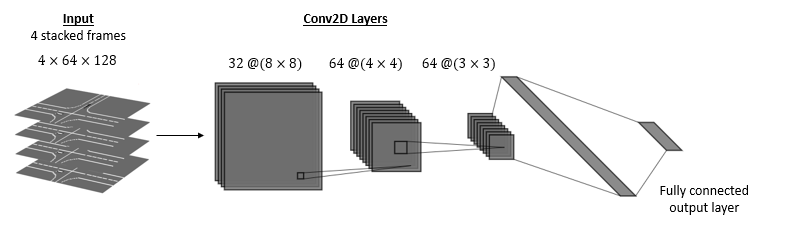}
    \caption{DQN: Convolutional Neural Network with the last 4 frames.}
    \label{fig:DQN Architecture}
\end{figure}

\textit{Highway-env} offers different types of observation spaces based on the complexity and information needs of the environment setup. In this study, the grayscale observation is used to represent the environment as a grayscale image (similar to a top-down view), where the positions of vehicles and road boundaries are visualized as pixel intensities.  In a driving scenario, it’s challenging to determine the speed or direction of other vehicles with just a
single snapshot. As a result, the four most recent frames are stacked together to capture temporal information, providing the model with a sense of motion and changes in the environment. This
stacked representation was fed into a Convolutional Neural Network (CNN) architecture within the DQN framework as depicted in Fig. \ref{fig:DQN Architecture}.
The agent must drive a vehicle by controlling its acceleration chosen from the set of abstract meta-actions in a predefined path from the starting point to the target point. To that end, the
following space of meta-actions are defined:
\begin{equation}
\mathcal{A}_i=[\texttt{Drive Slower, Maintain Speed (IDLE), Drive Faster}]
\end{equation}
Meta-actions are rather slow to affect the vehicle's state and are thus executed at a low frequency of 1 Hz.
This design choice simplifies the action space, which in turn causes the agent to concentrate its efforts on strategic speed adjustments rather than deal with adjusted control inputs. Executing actions at a lower frequency also makes the smoother vehicle behavior possible and improves the stability of the learning process.

\subsection{\textsc{RL With CLIP Feedback}}
\begin{figure}[!ht]
    \centering
    \includegraphics[width=0.75\linewidth]{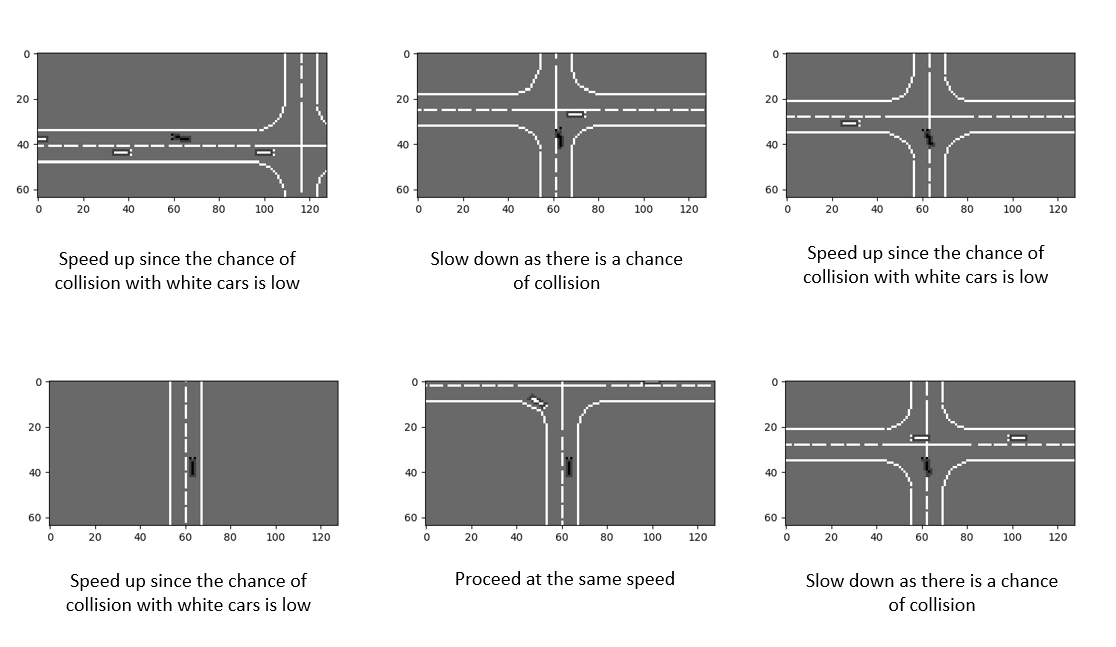}
    \caption{A summary of the (scenario, instruction) pairs in various driving situations at an unsignalized intersection, used for calibrating the CLIP model.}
    \label{fig:pairs}
\end{figure}

This section presents how VLMs, particularly CLIP, is used in this paper to improve the AV’s decision-making
process in various intersection scenarios. CLIP provides the model with associated visual inputs and natural language instructions, which can be used to guide the agent’s behavior based on
contextual understanding. The approach leverages transfer learning, which is particularly useful given the relatively small dataset available for refinement and modifications. Transfer learning facilitates the use of features learned by CLIP, which have been pre-trained on large, diverse datasets. Transferring these learned features to our dataset enables the model to improve performance, even with limited training data. This process avoids updating the entire model; most of the model’s weights are unchanged,
and only specific layers are refined, which causes faster training times and reduces GPU usage. To do that, a dataset including 500 images from different frames of the intersection scenarios was
collected. These images represented the diverse situations the AV could encounter during training. Alongside the images, 500 corresponding instructions were defined as single-sentence text prompts,
serving as labels for each image. These instructions are based on the vehicle’s current visual input. Fig. \ref{fig:pairs} illustrates the dataset’s structure used to optimize the CLIP model.
After adjusting CLIP, the trained model was integrated into the AV’s decision-making system. Hence, CLIP was used as a reward model
to train AV in an unsignalized intersection scenario. In each training step, the current environment state of the AV, which is represented by its visual input, is passed through the vision encoder of the CLIP model. Simultaneously, the three possible actions in the environment $\mathcal{A}$ passes again through the text encoder. These actions are encoded as text embeddings corresponding to each potential behavior the AV might adopt. By utilizing the enhanced CLIP model, the agent receives rewards from CLIP’s reward model whenever its action aligns with the instruction output by CLIP for the current scenario. This allows the agent to align its behavior with the predefined instructions and helps it make decisions that are consistent with the desired driving behaviors at the unsignalized intersection. To calculate the reward, the CLIP model calculates the cosine similarity between the visual representation of the current state and the text embedding of the corresponding natural language prompt, which describes the expected human behavior in a similar scenario as follows:
\begin{equation}
    R_{CLIP} = \frac{\mathcal{E}_{image} \cdot \mathcal{E}_{text}}{\|\mathcal{E}_{image}\| \|\mathcal{E}_{text}\|} 
\end{equation}
where $\mathcal{E}_{image}$ is the embedding vector from the image encoder, and $\mathcal{E}_{text}$ is the embedding vector generated by the text encoder. To ensure that the basic reward structure from the simulation environment does not dominate the training process, CLIP’s reward is multiplied by a weight determined during calibration and refinement. This scaling factor allows us to balance the influence of CLIP’s reward with the basic reward from the simulation.  The basic reward structure provided by the simulation environment includes three primary components as follows:
\vspace{0.3cm}
\begin{equation}
R_{Basic} = 
\begin{cases} 
      r_{\text{speed}} & \text{if the agent maintains an efficient speed,} \\
      r_{\text{collision}} & \text{if the ego-vehicle collides with another vehicle,} \\
      r_{\text{destination}} & \text{if the agent successfully reaches its destination,} \\
      0 & \text{otherwise.}
   \end{cases}
\end{equation}

\begin{figure}[!ht]
    \centering
    \includegraphics[width= 1 \linewidth]{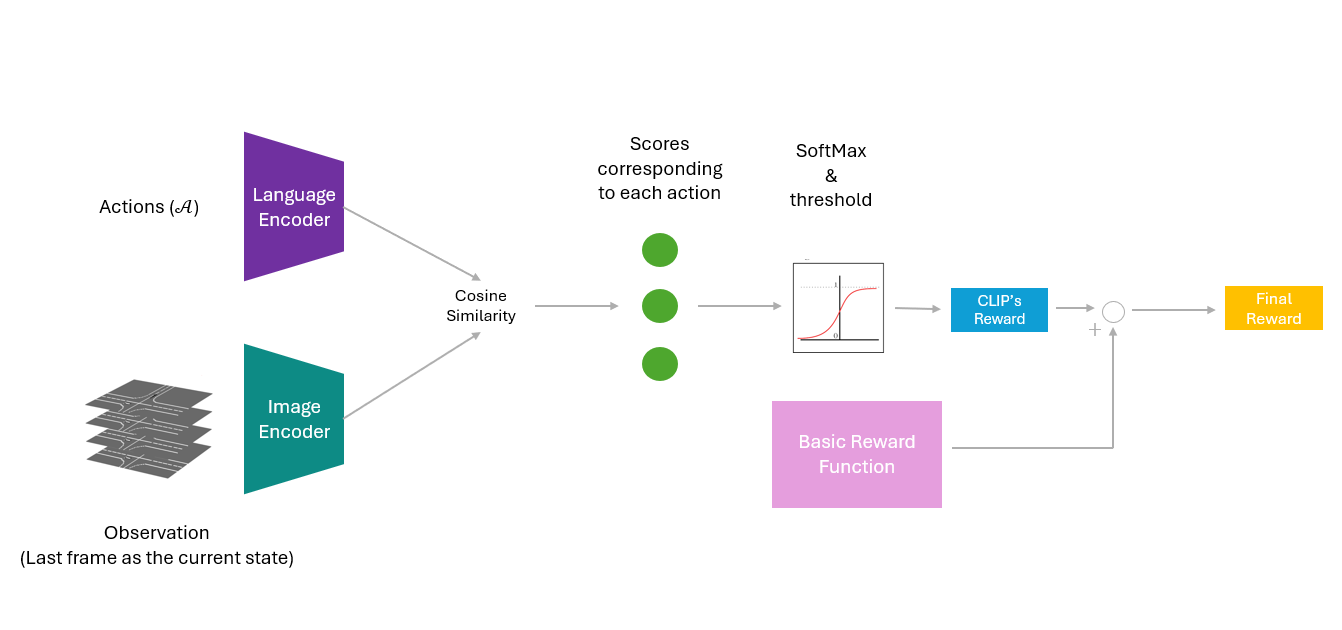}
    \caption{Architecture for CLIP as a reward model.}
    \label{fig:reward}
\end{figure}

\vspace{0.2cm}
where \( r_{\text{speed}} \in \mathbb{R}_+ \) represents the reward for maintaining an efficient speed while driving, \( r_{\text{collision}} \in \mathbb{R}_- \) denotes the penalty for a collision with another vehicle, and \( r_{\text{destination}} \in \mathbb{R}_+ \) is the reward for successfully reaching the destination. In this study, the values were chosen as \( r_{\text{speed}} = 1 \), \( r_{\text{collision}} = -5 \), and \( r_{\text{destination}} = 2 \). Finally, the weighted reward from CLIP is then combined with the reward from the basic reward structure provided by the simulation environment:
\begin{equation}
    R_{Final} = R_{Basic} + W_c  \times R_{CLIP} 
\end{equation}

where $W_c = 1.2$ is the corresponding weight, and $R_{CLIP}$ is the reward by CLIP reward model. Fig. \ref{fig:reward} shows how the reward mechanism using CLIP as a reward model works.

In this paper, DQN and PPO are employed as Deep Reinforcement Learning (DRL) methods while incorporating CLIP's reward model to guide the agent in optimizing its behavior in a human-like decision-making manner, at an unsignalized four-way intersection. Moreover, the performance of these two methods is compared with and without using CLIP and discusses the results regarding the safety and efficiency of the proposed framework. Algorithm 1 provides more detailed steps via pseudo-code for the DQN implementation. The simulations were conducted with a frequency of 15 Hz to ensure a realistic representation of dynamic traffic patterns. The computational experiments used an NVIDIA GeForce RTX 3050 Ti GPU and an AMD Ryzen 7-5800H CPU @3.20 GHz. The effectiveness of the PPO and DQN algorithms was influenced by the tuning of hyperparameters, which is demonstrated in Table \ref{tab:hyperparameters}. Also, to implement the RL algorithms in this work, we used the \texttt{Stable-Baselines3} \cite{raffin2021stable}.
\begin{table}[ht]
\centering
\caption{Hyperparameters used in DQN and PPO algorithms.}
\begin{tabularx}{\columnwidth}{Xcc}
\toprule
\textbf{Hyperparameter} & \textbf{DQN} & \textbf{PPO} \\
\midrule
$N_{steps}$           & 8000             & 8000  \\
Learning rate ($\alpha$)           & 0.0005             & 0.0005   \\
Discount factor ($\gamma)$        & 0.95               & 0.99      \\
Replay buffer size      & 15000              & -               \\
Batch size              & 32                 & 64               \\
Epsilon decay           & Linear              & -                \\
Epochs                  & -                 & 10                 \\
Clip range              & -                & 0.2                 \\
$\lambda$               & -                & 0.95                \\
\bottomrule
\end{tabularx}
\vspace{0.2cm}
\label{tab:hyperparameters}
\end{table}
In our implementation of CLIP, the Vision Transformer (ViT) architecture \cite{dosovitskiy2020image} is utilized as the image encoder to process visual inputs. The specific hyperparameters for this setup are detailed in Table \ref{tab:clip_hyperparameters} in the Appendix.

\begin{algorithm}[!ht]
\caption{Pseudo-code of the proposed CLIP-Based reward function with DQN}
\begin{algorithmic}[1]
\State Initialize the $Q$-network $Q(s, a; \theta)$ with random weights $\theta$
\State Initialize the target $Q$-network $\hat{Q}(s, a; \theta^{-})$ with $\theta^{-} \gets \theta$
\State Initialize the replay buffer $B$ with capacity $N$
\State Set hyperparameters: learning rate $\alpha$, discount factor $\gamma$, epsilon $\epsilon$, batch size $N_b$
\For{each episode}
    \State Initialize state $s_0$
    \For{each step in the episode}
        \State With probability $\epsilon$, select a random action $a_t$, otherwise select:
        \[a_t = \arg\max_a Q(s_t, a; \theta)\]
        \State Execute action $a_t$ and observe reward $r_b$ and next state $s_{t+1}$
        \State Pass the visual observation to the CLIP model to get the instruction
        \State Comparing the taken action with suggested action based on CLIP 
        \If{$a_t$ =  CLIP instruction}
        \State \[ r_f = r_b + w \times r_c\]
        \Else
        \State Use only the basic reward $r_f = r_b$
        \EndIf
        \State Store the transition $(s_t, a_t, r_f, s_{t+1})$ in the replay buffer $B$
        \State Sample a random mini-batch of transitions $(s_j, a_j, r_j, s_{j+1})$ from the buffer $B$
        \State Set the target for each transition:
        \[
        y_j = r_j + \gamma \max_{a'} \hat{Q}(s_{j+1}, a'; \theta^{-})
        \]
        \State Perform a gradient descent step on:
        \[
        L(\theta) = \frac{1}{N_b} \sum_{j=1}^{N_b} \left( Q(s_j, a_j; \theta) - y_j \right)^2
        \]
        \State Update the target network: $\hat{Q}(s, a; \theta^{-}) \gets Q(s, a; \theta)$ every $C$ steps
    \EndFor
\EndFor
\label{alg:dqn}
\end{algorithmic}
\end{algorithm}

\section{\textsc{Experiments}}
\label{s3}
In this section, the performance of the proposed framework is evaluated using DQN and PPO as RL algorithms and CLIP as a VLM to build a reward model. An overview of the experimental setup is also provided, including the environment configurations, hyperparameters, and training protocols used for DQN and PPO. Subsequently, the results are presented based on several evaluation metrics to compare the performance of the two RL algorithms with and without the CLIP-based reward function. In addition, the influence of CLIP on agent behavior at unsignalized intersections will be evaluated based on key factors such as collision rate, success rate, and alignment with human-like decision-making behaviors through a series of simulation experiments.

\subsection{\textsc{Training Evaluation}}
Training sessions on environmental configuration were conducted, as presented in the Table. \ref{tab:config} in the Appendix. The simulations were conducted for 8,000 steps to train the proposed models using DQN and PPO algorithms separately. The aim was to observe and analyze the behavior of the AVs in an unsignalized four-way intersection. Several key factors were measured, including the success rate, collision rate, and timeout rate (instances when an the AV fails to reach its destination). The primary objective of these simulations was to assess the overall generalization ability of the trained policies with CLIP used as a reward model.
\begin{figure}[!ht]
    \centering
    \begin{subfigure}[b]{0.45\linewidth}
        \centering
        \includegraphics[width=\linewidth]{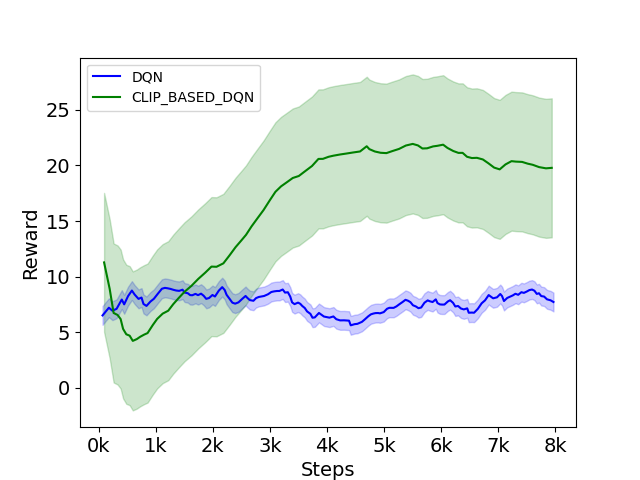}
        \caption{DQN}
        \label{fig:enter-label-left}
    \end{subfigure}
    \hfill
    \begin{subfigure}[b]{0.45\linewidth}
        \centering
        \includegraphics[width=\linewidth]{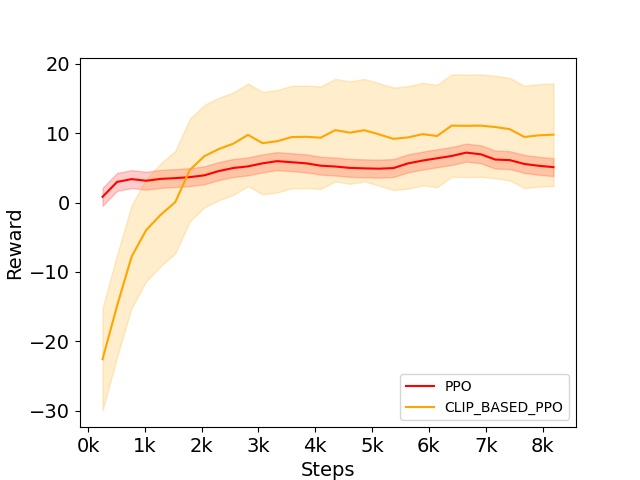}
        \caption{PPO}
        \label{fig:enter-label-right}
    \end{subfigure}
    \caption{The change of average reward per episode through the training process for CLIP-based and non-CLIP-based DQN, and PPO.}
    \label{fig:training-reward}
\end{figure}

\begin{figure}[!ht]
    \centering
    \begin{subfigure}[b]{0.45\linewidth}
        \centering
        \includegraphics[width=\linewidth]{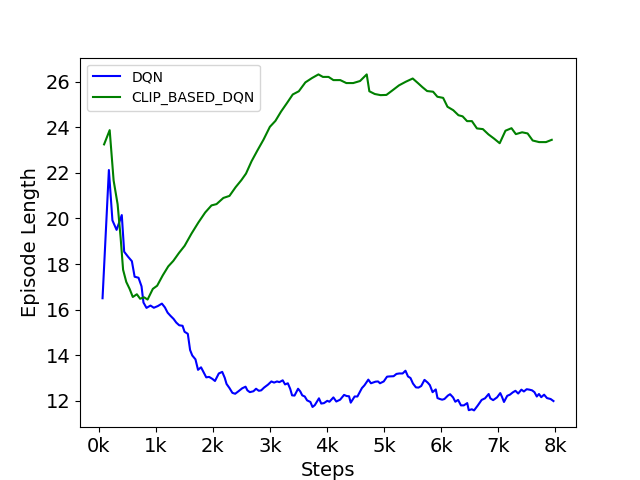}
        \caption{DQN}
        \label{fig:enter-label-left}
    \end{subfigure}
    \hfill
    \begin{subfigure}[b]{0.45\linewidth}
        \centering
        \includegraphics[width=\linewidth]{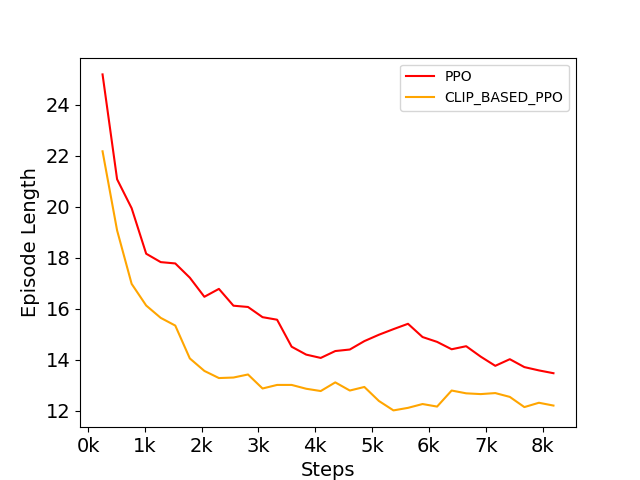}
        \caption{PPO}
        \label{fig:enter-label-right}
    \end{subfigure}
    \caption{The change of episode's length through the training process for CLIP-based and non-CLIP-based DQN, and PPO.}
    \label{fig:Episode length}
\end{figure}
During the experiments, each episode’s average reward and duration were recorded and assessed or the agent to evaluate the performance of the training session under each method. Both the episodic reward and the episode’s length were evaluated, as shown in Figs. \ref{fig:training-reward}, and \ref{fig:Episode length}, respectively.

This indicates that integrating the CLIP-based reward model considerably enhances the performance of both DQN and PPO algorithms. The CLIP-based DQN shows better performance than the CLIP-based PPO, which also indicates that rewards increase steadily over time. In contrast, both DQN and PPO without CLIP show limited learning progress, as their reward variations do not show a meaningful increase. This shows that the agent struggles to optimize their behavior effectively in such a complex driving scenario. While CLIP-based PPO also performs better than standard PPO, the improvement is more pronounced in CLIP-based DQN, which emphasizes the greater impact of the CLIP reward model when applied to DQN. Based on Fig. \ref{fig:Episode length}, a lower episode length suggests that the agent may have encountered a collision with other vehicles, which caused the episode to terminate prematurely. On the other hand, since the episode duration is set to 30 seconds, a high episode length indicates that the agent failed to complete the left-turn task within the allocated time. In contrast, for the CLIP-based DQN, the episode length gradually converges to an optimal value over the training process. This suggests that the agent learns to reduce its speed while approaching the intersection to avoid collision and successfully completes its task of turning left.

\subsection{\textsc{Post-training Evaluation}}
The post-training evaluation involves deploying the trained agent utilizing CLIP-based DQN, CLIP-based PPO, vanilla DQN, and PPO in an intersection setting to assess its navigation abilities across 100 testing episodes. The trained agent employs its learned policy to make real-time decisions at every time step. The agent starts in an initial state and chooses the best action based on its policy. Following this, it receives visual observation and reward and transitions to the next state. This iterative process goes on until the agent arrives at a terminal state during the episode. 

Algorithm 2 in the Appendix shows how the required steps to do the evaluation process. In the proposed scheme, after initialization, the trained agent is assigned an action \(a_t\) according to its policy \(\pi_{\mathcal{A}}\). Subsequently, the agent executes this action, receives a reward \(r_t\), and observes the next state. The evaluation continues until the episode terminates, either due to a collision, successful arrival at the destination, or truncation.
If a collision occurs during the episode, the collision counter \(c\) is incremented by one, and the episode is terminated.
In contrast, if the agent successfully reaches its destination, the success counter \(s\) is incremented by one. During the evaluation, the agent’s speed at each time step is appended to a list \(v\), which provides the possibility for calculating the average speed \(\bar{v}\) at the end of the evaluation process. At the end of all episodes, the algorithm outputs the normalized metrics, which also facilitates a generalized assessment of the trained agent's behavior across multiple scenarios. 

\begin{table}[ht]
\centering
\caption{Comparison of performance for different methods across scenarios with varying initial numbers of on-road vehicles.}
\begin{tabular}{lcccc}
\toprule
\textbf{Method} & \textbf{Vehicles} & \textbf{Success (\%)} & \textbf{Collision (\%)} & \textbf{Time-out (\%)} \\
\midrule
\multirow{3}{*}{CLIP-DQN} & $1$  & \textbf{96} & \textbf{4} & 0 \\
                          & $3$  & \textbf{84} & \textbf{16} & 0 \\
                          & $6$  & \textbf{72} & \textbf{28} & 0 \\
\midrule
\multirow{3}{*}{CLIP-PPO} & $1$  & 38 & 8 & 54 \\
                          & $3$  & 38 & 20 & 42 \\
                          & $6$  & 26 & 26 & 48 \\
\midrule
\multirow{3}{*}{DQN}      & $1$  & 70 & 30 & 0 \\
                          & $3$  & 56 & 44 & 0 \\
                          & $6$  & 50 & 50 & 0 \\
\midrule
\multirow{3}{*}{PPO}      & $1$  & 78 & 22 & 0 \\
                          & $3$  & 60 & 40 & 0 \\
                          & $6$  & 56 & 44 & 0 \\
\bottomrule
\end{tabular}
\vspace{0.2cm}
\label{tab:comparison}
\end{table}

Table \ref{tab:comparison} summarizes the success, collision, and time-out rates during the testing process with different trained policies. We extend our analysis to scenarios involving 1, 3, and 6 HVs to examine the impact of vehicle count on the performance of CLIP-DQN, CLIP-PPO, DQN, and PPO. It is observed that the success rate, collision rate, and time-out rate for CLIP-DQN in the case with 1 on-road vehicle are 96\%, 4\%, and 0\%, respectively. These values are 38\%, 8\%, and 54\% for CLIP-PPO, 70\%, 30\%, and 0\% for DQN, and 78\%, 22\%, and 0\% for PPO. 
Similarly, in scenarios with 3 and 6 HVs, CLIP-DQN achieves the best success and collision rates, outperforming the other methods. However, as the number of on-road vehicles increases, the success rate decreases, and the collision rate rises due to the increasing complexity of the environment. For CLIP-DQN, the success rate decreases from 96\% to 84\% and then to 72\%, while the collision rate increases from 4\% to 16\% and then to 28\%, respectively. One notable drawback of CLIP-PPO is its non-smooth movement, often preventing the agent from completing the turning task within the specified time. This limitation is reflected in its significantly higher time-out rate compared to other methods.
\begin{figure}[ht]
    \centering
    \begin{subfigure}[b]{0.49\linewidth}
        \centering
        \includegraphics[width=\linewidth]{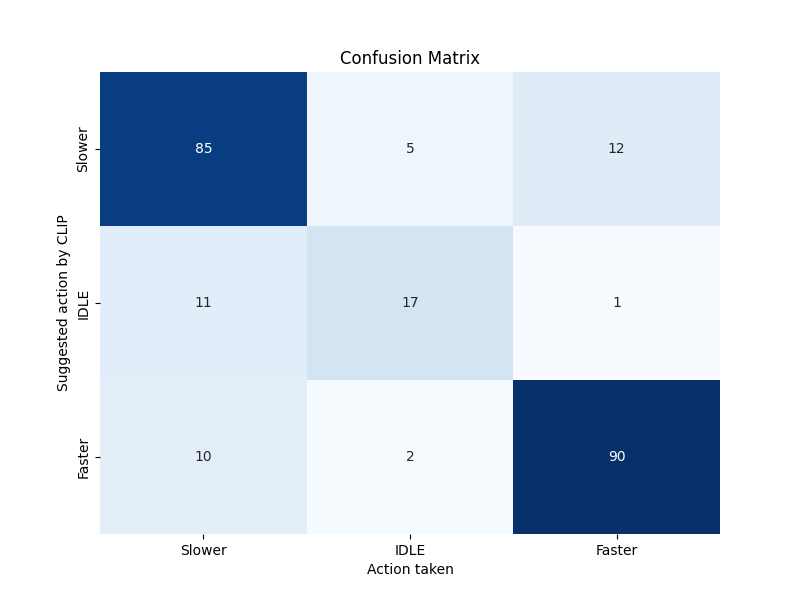}
        \caption{CLIP-based DQN}
        \label{fig:enter-label-left}
    \end{subfigure}
    \hfill
    \begin{subfigure}[b]{0.49\linewidth}
        \centering
        \includegraphics[width=\linewidth]{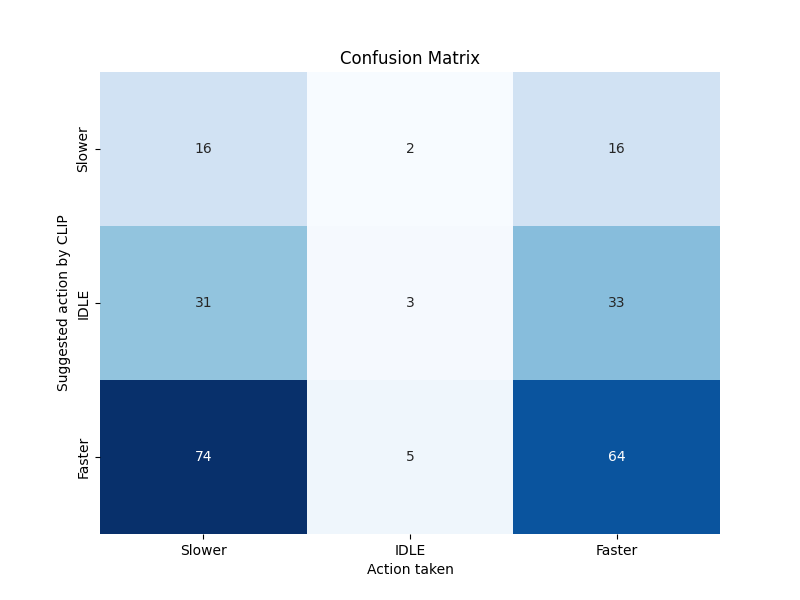}
        \caption{CLIP-based PPO}
        \label{fig:enter-label-right}
    \end{subfigure}
    \caption{Confusion matrices comparing the actions suggested by the CLIP model and those taken by the agent.}
    \label{fig:confusion_matrices}
\end{figure}
Confusion matrices in Fig. \ref{fig:confusion_matrices} illustrate the alignment between the actions taken by the trained agents using CLIP-based DQN and PPO
and the actions suggested by the CLIP model. In these matrices, the rows represent the actions suggested by CLIP, and the columns represent the actions actually taken by the agent. For the agent trained with CLIP-based DQN, the actions taken are generally well-aligned with the CLIP recommendations. Most of the time, the agent follows the suggested actions, which shows the satisfactory coordination between the model and the agent’s decisions. In contrast, for the agent trained using CLIP-based PPO, there is a significant misalignment between the suggested and actual actions, particularly when it comes to slowing down. This misalignment is most evident in the agent's failure to decelerate when approaching the intersection, which results in higher collision rates.

\begin{figure}[!ht]
    \centering
    \includegraphics[width=0.7\linewidth]{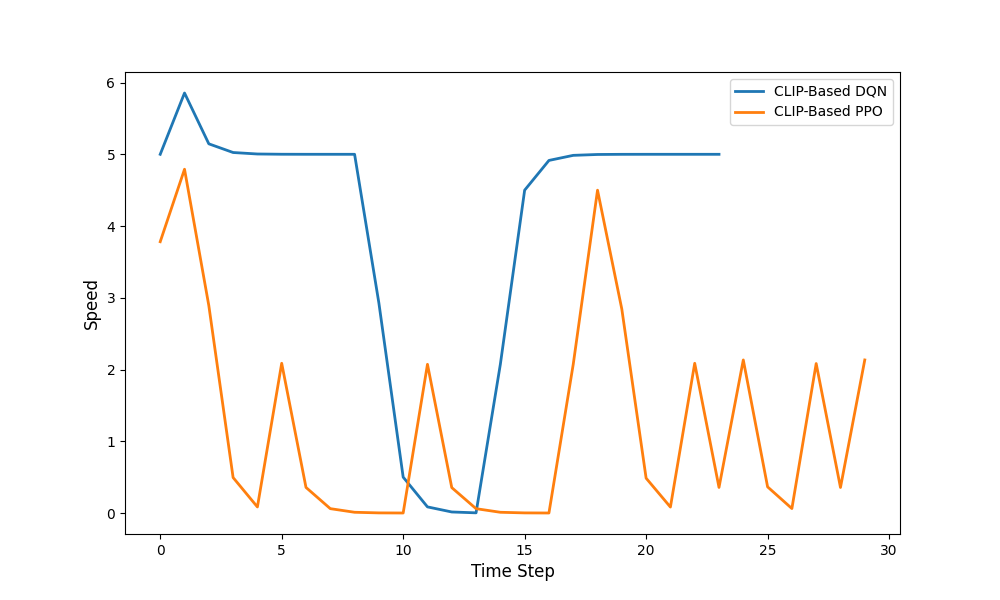}
    \caption{Ego vehicle speed's profile over time.}
    \label{fig:speed}
\end{figure}
The speed profile in Fig. \ref{fig:speed} demonstrates the behavior of the AV over time during the test. With CLIP-based DQN, the speed profile follows an anticipated pattern. As the AV approaches the intersection, its speed to brake reduces. Once the intersection/road is clear to make the left turn, the AV accelerates to complete the maneuver. This gradual shift between slowing down and speeding up also confirms effective decision-making in response to the environment. On the other side, the speed profile for CLIP-based PPO shows significant fluctuations. The AV does not exhibit smooth driving and struggles to maintain appropriate speed during critical phases, particularly when approaching the intersection. These fluctuations suggest that the CLIP-based PPO agent is less consistent in following the desired behavior, which leads to less optimal and potentially unsafe driving actions for other road users as well.

\begin{figure}[ht]
    \centering
    \begin{subfigure}[b]{0.49\linewidth}
        \centering
        \includegraphics[width=\linewidth]{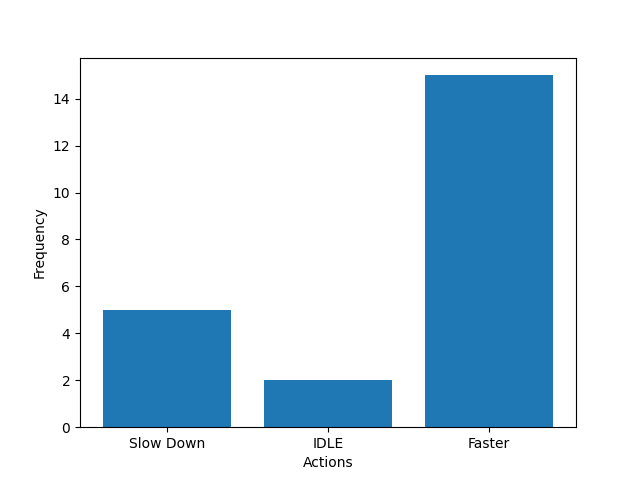}
        \caption{}
        \label{fig:freq}
    \end{subfigure}
    \hfill
    \begin{subfigure}[b]{0.49\linewidth}
        \centering
        \includegraphics[width=\linewidth]{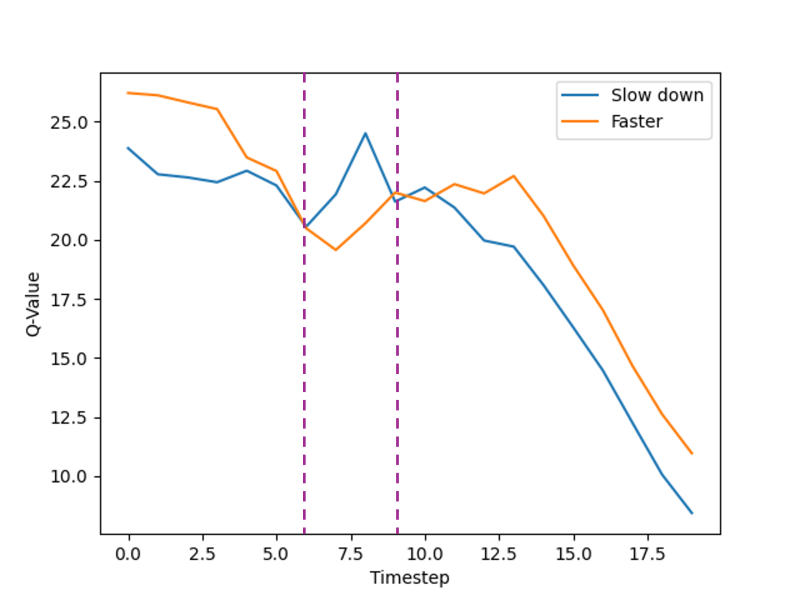}
        \caption{}
        \label{fig:qs}
    \end{subfigure}
    \caption{a) Frequency of each action with CLIP-based DQN b) Q-Values from CLIP-based DQN, for each action over time.}
    \label{fig:figs}
\end{figure}

Figure \ref{fig:freq} shows the frequency distribution of actions taken by the AV over multiple test episodes. The action \textit{Faster} was the most frequently selected by the agent, which also suggests the vehicle's preference in prioritizing increased speed. On the other hand, the action \textit{Slow Down} was chosen less frequently, and action \textit{IDLE} had the lowest frequency. One main goal is to complete the task as quickly as possible to improve traffic efficiency by reducing waiting times for other vehicles. At the same time, the policy is designed to help improve safety by avoiding collisions. The higher frequency of the action \textit{Faster} means that the trained policy has effectively learned to balance these objectives while minimizing unnecessary deceleration. Hence, the action \textit{IDLE} is less essential for the agent compared to other actions to accomplish these objectives.

Fig. \ref{fig:qs} shows the \textit{Q}-values associated with each action. At the beginning of the episode, the Q-values for the \textit{Faster} action are higher than those for \textit{Slow Down}, which shows that the AV prioritizes increasing its speed during the initial stages. As the AV approaches the intersection, the \textit{Q}-values for \textit{Slow Down} increase, which shows that the policy identifies the need to decelerate to navigate the intersection safely.

This also illustrates that the policy appropriately identifies caution in potentially risky scenarios,
which is also followed by the safety objectives of the task. After passing the intersection, the \textit{Q}-values for \textit{Faster} rise again, which indicates a shift back to prioritizing speed for efficient task completion.

\section{\textsc{Conclusion}}
\label{s4}
In this paper, a novel method was proposed to integrate VLMs, specifically CLIP, with RL algorithms such as PPO and DQN to improve the decision-making capabilities of AVs navigating through unsignalized four-way intersections, which is a complex environment to find the optimal policy with traditional approaches. The obtained results in this paper demonstrated by introducing a reward mechanism based on VLMs, the AV is facilitated to align its actions more closely with human-like decision-making patterns to address the limitations of rule-based systems. In addition, the conducted simulation experiments demonstrated that the agent achieved higher rewards when guided by this enhanced reward function, compared to using only the conventional reward function. This paper emphasizes the prospects of combining VLMs with RL techniques to develop more adaptive and context-aware AV systems. 
Despite the promising outcomes, future research needs to improve its evaluations of more generalized urban driving conditions with a more extended dataset. 
Furthermore, a human-in-the-loop framework could be developed to enhance the realism of the evaluation process. By including virtual reality environments and immediate human interaction, subjective human factor analysis could be performed to better understand how humans perceive and respond to AV behavior. This approach would allow for the refinement and modification of algorithms based on direct feedback for other practical applications.

\bibliographystyle{unsrt}  
\bibliography{references}  

\begin{thebibliography}{10}

\bibitem{huang2021driving}
Zhiyu Huang, Jingda Wu, and Chen Lv.
\newblock Driving behavior modeling using naturalistic human driving data with inverse reinforcement learning.
\newblock {\em IEEE transactions on intelligent transportation systems}, 23(8):10239--10251, 2021.

\bibitem{taghavifar}
Victor Rasidescu and Hamid Taghavifar.
\newblock Socially intelligent path-planning for autonomous vehicles using type-2 fuzzy estimated social psychology models.
\newblock {\em IEEE Access}, 2024.

\bibitem{cui2023drivellm}
Yaodong Cui, Shucheng Huang, Jiaming Zhong, Zhenan Liu, Yutong Wang, Chen Sun, Bai Li, Xiao Wang, and Amir Khajepour.
\newblock Drivellm: Charting the path toward full autonomous driving with large language models.
\newblock {\em IEEE Transactions on Intelligent Vehicles}, 2023.

\bibitem{taghavifar2024socially}
Hamid Taghavifar, Chongfeng Wei, and Leyla Taghavifar.
\newblock Socially intelligent reinforcement learning for optimal automated vehicle control in traffic scenarios.
\newblock {\em IEEE Transactions on Automation Science and Engineering}, 2024.

\bibitem{toghi2022social}
Behrad Toghi, Rodolfo Valiente, Dorsa Sadigh, Ramtin Pedarsani, and Yaser~P Fallah.
\newblock Social coordination and altruism in autonomous driving.
\newblock {\em IEEE Transactions on Intelligent Transportation Systems}, 23(12):24791--24804, 2022.

\bibitem{tram2019learning}
Tommy Tram, Ivo Batkovic, Mohammad Ali, and Jonas Sj{\"o}berg.
\newblock Learning when to drive in intersections by combining reinforcement learning and model predictive control.
\newblock In {\em 2019 IEEE Intelligent Transportation Systems Conference (ITSC)}, pages 3263--3268. IEEE, 2019.

\bibitem{ronecker2019deep}
Max~Peter Ronecker and Yuan Zhu.
\newblock Deep q-network based decision making for autonomous driving.
\newblock In {\em 2019 3rd international conference on robotics and automation sciences (ICRAS)}, pages 154--160. IEEE, 2019.

\bibitem{li2024large}
Yun Li, Kai Katsumata, Ehsan Javanmardi, and Manabu Tsukada.
\newblock Large language models for human-like autonomous driving: A survey.
\newblock {\em arXiv preprint arXiv:2407.19280}, 2024.

\bibitem{llms}
Mengyao Wu, F~Richard Yu, Peter~Xiaoping Liu, and Ying He.
\newblock Facilitating autonomous driving tasks with large language models.
\newblock {\em IEEE Intelligent Systems}, 2024.

\bibitem{li2023towards}
Xin Li, Yeqi Bai, Pinlong Cai, Licheng Wen, Daocheng Fu, Bo~Zhang, Xuemeng Yang, Xinyu Cai, Tao Ma, Jianfei Guo, et~al.
\newblock Towards knowledge-driven autonomous driving.
\newblock {\em arXiv preprint arXiv:2312.04316}, 2023.

\bibitem{zheng2024planagent}
Yupeng Zheng, Zebin Xing, Qichao Zhang, Bu~Jin, Pengfei Li, Yuhang Zheng, Zhongpu Xia, Kun Zhan, Xianpeng Lang, Yaran Chen, et~al.
\newblock Planagent: A multi-modal large language agent for closed-loop vehicle motion planning.
\newblock {\em arXiv preprint arXiv:2406.01587}, 2024.

\bibitem{ardashir}
Hamid Taghavifar and Ardashir Mohammadzadeh.
\newblock Integrating deep reinforcement learning and social-behavioral cues: A new human-centric cyber-physical approach in automated vehicle decision-making.
\newblock {\em Proceedings of the Institution of Mechanical Engineers, Part D: Journal of Automobile Engineering}, page 09544070241230126, 2024.

\bibitem{kiran2021deep}
B~Ravi Kiran, Ibrahim Sobh, Victor Talpaert, Patrick Mannion, Ahmad~A Al~Sallab, Senthil Yogamani, and Patrick P{\'e}rez.
\newblock Deep reinforcement learning for autonomous driving: A survey.
\newblock {\em IEEE Transactions on Intelligent Transportation Systems}, 23(6):4909--4926, 2021.

\bibitem{kwon2023reward}
Minae Kwon, Sang~Michael Xie, Kalesha Bullard, and Dorsa Sadigh.
\newblock Reward design with language models.
\newblock {\em arXiv preprint arXiv:2303.00001}, 2023.

\bibitem{chan2024dense}
Alex~J Chan, Hao Sun, Samuel Holt, and Mihaela van~der Schaar.
\newblock Dense reward for free in reinforcement learning from human feedback.
\newblock {\em arXiv preprint arXiv:2402.00782}, 2024.

\bibitem{randlov1998learning}
Jette Randl{\o}v and Preben Alstr{\o}m.
\newblock Learning to drive a bicycle using reinforcement learning and shaping.
\newblock In {\em ICML}, volume~98, pages 463--471, 1998.

\bibitem{bai2022constitutional}
Yuntao Bai, Saurav Kadavath, Sandipan Kundu, Amanda Askell, Jackson Kernion, Andy Jones, Anna Chen, Anna Goldie, Azalia Mirhoseini, Cameron McKinnon, et~al.
\newblock Constitutional ai: Harmlessness from ai feedback.
\newblock {\em arXiv preprint arXiv:2212.08073}, 2022.

\bibitem{scheurer2023training}
J{\'e}r{\'e}my Scheurer, Jon~Ander Campos, Tomasz Korbak, Jun~Shern Chan, Angelica Chen, Kyunghyun Cho, and Ethan Perez.
\newblock Training language models with language feedback at scale.
\newblock {\em arXiv preprint arXiv:2303.16755}, 2023.

\bibitem{xie2023text2reward}
Tianbao Xie, Siheng Zhao, Chen~Henry Wu, Yitao Liu, Qian Luo, Victor Zhong, Yanchao Yang, and Tao Yu.
\newblock Text2reward: Automated dense reward function generation for reinforcement learning.
\newblock {\em arXiv preprint arXiv:2309.11489}, 2023.

\bibitem{devlin2018bert}
Jacob Devlin.
\newblock Bert: Pre-training of deep bidirectional transformers for language understanding.
\newblock {\em arXiv preprint arXiv:1810.04805}, 2018.

\bibitem{achiam2023gpt}
Josh Achiam, Steven Adler, Sandhini Agarwal, Lama Ahmad, Ilge Akkaya, Florencia~Leoni Aleman, Diogo Almeida, Janko Altenschmidt, Sam Altman, Shyamal Anadkat, et~al.
\newblock Gpt-4 technical report.
\newblock {\em arXiv preprint arXiv:2303.08774}, 2023.

\bibitem{touvron2023llama}
Hugo Touvron, Thibaut Lavril, Gautier Izacard, Xavier Martinet, Marie-Anne Lachaux, Timoth{\'e}e Lacroix, Baptiste Rozi{\`e}re, Naman Goyal, Eric Hambro, Faisal Azhar, et~al.
\newblock Llama: Open and efficient foundation language models.
\newblock {\em arXiv preprint arXiv:2302.13971}, 2023.

\bibitem{NIPS2017_d5e2c0ad}
Paul~F Christiano, Jan Leike, Tom Brown, Miljan Martic, Shane Legg, and Dario Amodei.
\newblock Deep reinforcement learning from human preferences.
\newblock In I.~Guyon, U.~Von Luxburg, S.~Bengio, H.~Wallach, R.~Fergus, S.~Vishwanathan, and R.~Garnett, editors, {\em Advances in Neural Information Processing Systems}, volume~30. Curran Associates, Inc., 2017.

\bibitem{wang2024rl}
Yufei Wang, Zhanyi Sun, Jesse Zhang, Zhou Xian, Erdem Biyik, David Held, and Zackory Erickson.
\newblock Rl-vlm-f: Reinforcement learning from vision language foundation model feedback.
\newblock {\em arXiv preprint arXiv:2402.03681}, 2024.

\bibitem{yu2023language}
Wenhao Yu, Nimrod Gileadi, Chuyuan Fu, Sean Kirmani, Kuang-Huei Lee, Montse~Gonzalez Arenas, Hao-Tien~Lewis Chiang, Tom Erez, Leonard Hasenclever, Jan Humplik, et~al.
\newblock Language to rewards for robotic skill synthesis.
\newblock {\em arXiv preprint arXiv:2306.08647}, 2023.

\bibitem{mao2023gpt}
Jiageng Mao, Yuxi Qian, Junjie Ye, Hang Zhao, and Yue Wang.
\newblock Gpt-driver: Learning to drive with gpt.
\newblock {\em arXiv preprint arXiv:2310.01415}, 2023.

\bibitem{sadigh2}
Minae Kwon, Sang~Michael Xie, Kalesha Bullard, and Dorsa Sadigh.
\newblock Reward design with language models.
\newblock {\em arXiv preprint arXiv:2303.00001}, 2023.

\bibitem{rocamonde2023vision}
Juan Rocamonde, Victoriano Montesinos, Elvis Nava, Ethan Perez, and David Lindner.
\newblock Vision-language models are zero-shot reward models for reinforcement learning.
\newblock {\em arXiv preprint arXiv:2310.12921}, 2023.

\bibitem{adeniji2023language}
Ademi Adeniji, Amber Xie, Carmelo Sferrazza, Younggyo Seo, Stephen James, and Pieter Abbeel.
\newblock Language reward modulation for pretraining reinforcement learning.
\newblock {\em arXiv preprint arXiv:2308.12270}, 2023.

\bibitem{radford2021learning}
Alec Radford, Jong~Wook Kim, Chris Hallacy, Aditya Ramesh, Gabriel Goh, Sandhini Agarwal, Girish Sastry, Amanda Askell, Pamela Mishkin, Jack Clark, et~al.
\newblock Learning transferable visual models from natural language supervision.
\newblock In {\em International conference on machine learning}, pages 8748--8763. PMLR, 2021.

\bibitem{highway-env}
Edouard Leurent.
\newblock An environment for autonomous driving decision-making.
\newblock \url{https://github.com/eleurent/highway-env}, 2018.

\bibitem{brockman2016openai}
G~Brockman.
\newblock Openai gym.
\newblock {\em arXiv preprint arXiv:1606.01540}, 2016.

\bibitem{mnih2013playing}
Volodymyr Mnih.
\newblock Playing atari with drl.
\newblock {\em arXiv preprint arXiv:1312.5602}, 2013.

\bibitem{schulman2017proximal}
John Schulman, Filip Wolski, Prafulla Dhariwal, Alec Radford, and Oleg Klimov.
\newblock Ppo algorithms.
\newblock {\em arXiv preprint arXiv:1707.06347}, 2017.

\bibitem{schulman2015trust}
John Schulman.
\newblock Trust region policy optimization.
\newblock {\em arXiv preprint arXiv:1502.05477}, 2015.

\bibitem{raffin2021stable}
Antonin Raffin, Ashley Hill, Adam Gleave, Anssi Kanervisto, Maximilian Ernestus, and Noah Dormann.
\newblock Stable-baselines3: Reliable reinforcement learning implementations.
\newblock {\em Journal of Machine Learning Research}, 22(268):1--8, 2021.

\bibitem{dosovitskiy2020image}
Alexey Dosovitskiy.
\newblock An image is worth 16x16 words: Transformers for image recognition at scale.
\newblock {\em arXiv preprint arXiv:2010.11929}, 2020.

\bibitem{kingma2014adam}
Diederik~P Kingma.
\newblock Adam: A method for stochastic optimization.
\newblock {\em arXiv preprint arXiv:1412.6980}, 2014.

\end{thebibliography}

\clearpage
\appendix
\section{Appendix}
\subsection{Trajectories}
\begin{figure}[H]
    \centering
    \includegraphics[width=0.75\linewidth]{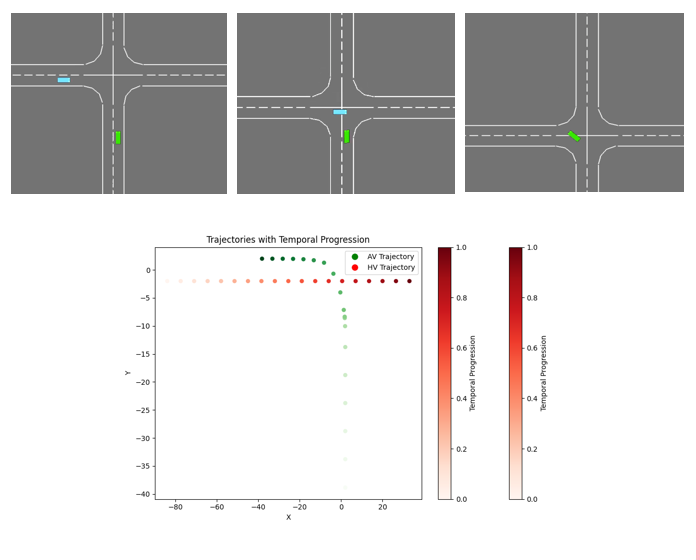}
    \caption{Scenario 1, the AV yields to the HV approaching from the west.}
    \label{fig:scenario 1}
\end{figure}

\begin{figure}[!ht]
    \centering
    \includegraphics[width=0.75\linewidth]{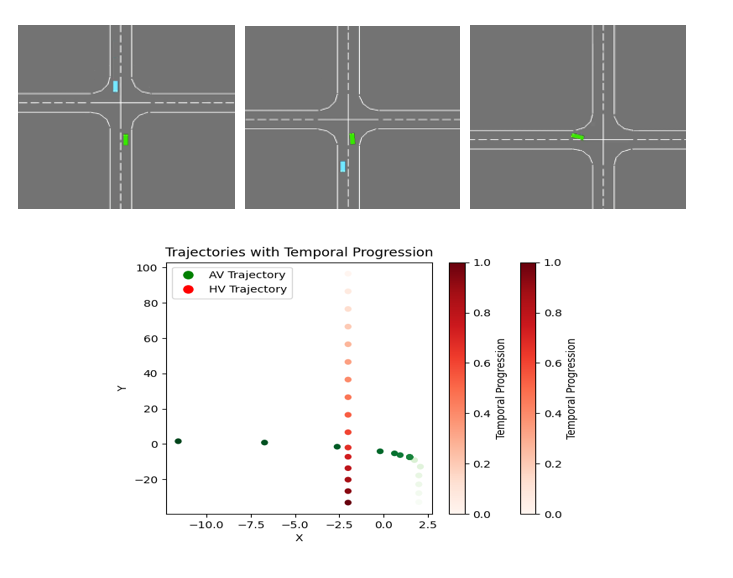}
    \caption{Scenario 2, the AV yields to the HV approaching from the north.}
    \label{fig:scenario 2}
\end{figure}

\begin{figure}[H]
    \centering
    \includegraphics[width=0.75\linewidth]{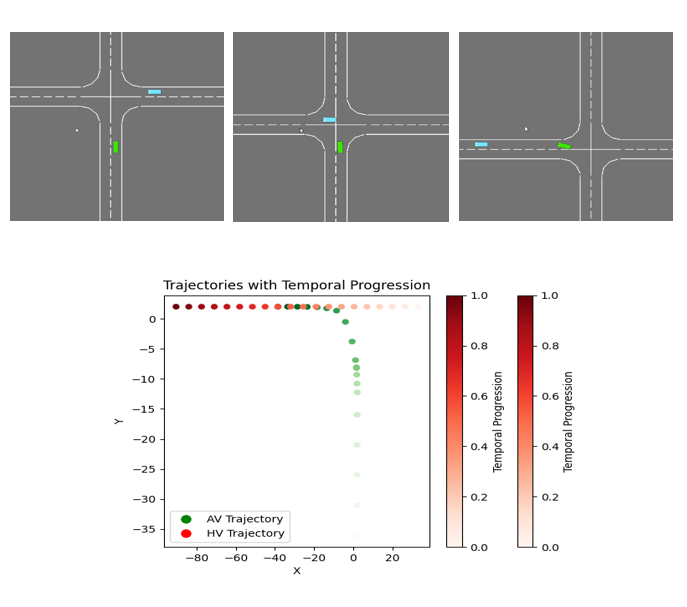}
    \caption{Scenario 3, the AV yields to the HV approaching from the east.}
    \label{fig:scenario 3}
\end{figure}

Figures \ref{fig:scenario 1}, \ref{fig:scenario 2}, and \ref{fig:scenario 3} depict the trajectories of the AV and HVs in three intersection scenarios using an agent trained with the CLIP-based DQN. In each scenario, the AV starts from the same initial position (south), while HVs approach the intersection from different directions. These representations demonstrate the AV's decision-making process and emphasize its altruistic behavior. Also, the bottom component in each representation corresponds to the temporal progression of the trajectories and the intensity of the gradient shows the temporal progress of the vehicles. In Fig. \ref{fig:scenario 1}, the AV begins its trajectory from the south side of the intersection, and an HV approaches from the east. The AV slows down to allow the HV to pass safely through the intersection. Once the intersection is clear, the AV accelerates and executes a left turn. In Fig. \ref{fig:scenario 2}, and \ref{fig:scenario 3}, the HV starts from north, and west, respectively. Similar to Scenario 1, the AV yields to the HV to maintain a cautious approach, and then it speeds up and completes the left turn.


\subsection{CLIP Implementation Details}
Different image encoder architectures are available for CLIP. In this paper, the experiments are conducted with the ViT. Specifically, the ViT-B/32 model on our dataset for 15 epochs was modified using an Adam optimizer \cite{kingma2014adam}. For the purpose of transfer learning, the gradients were updated only for the
last layer of the model and left the earlier layers frozen. The hyperparameters used in this setup are listed in Table \ref{tab:clip_hyperparameters}.

\begin{table}[H]
  \centering
    \caption{CLIP ViT-B/32 hyperparameters.}
  \resizebox{\textwidth}{!}{
    \begin{tabular}{
      l
      S
      S
      S
      S
      S
      S
      S
      S
      S
    }
    \toprule
    & \text{Learning} & \text{Embedding} & \text{Input}& \multicolumn{3}{c}{Vision Transformer} & \multicolumn{3}{c}{Text Transformer} \\
    \cmidrule(lr){5-7} \cmidrule(lr){8-10}
    {Model} & {rate} & {dimension} & {resolution} & {layers} & {width} & {heads} & {layers} & {width} & {heads} \\
    \midrule
    ViT-B/32 & 5 \(\times 10 ^ {-4}\) & 512 & 224 & 12 & 768 & 12 & 12 & 512 & 8 \\
    \bottomrule
    \end{tabular}
  }
  \captionsetup{font=small, justification=centering}
  \label{tab:clip_hyperparameters}
\end{table}

\subsection{Simulation Environment}
The environment is intersection-v1 from \textit{highway-env} with discrete actions (i.e., speed up or slow down) and continuous observations (i.e., grayscale frames). To train the agents, the \texttt{Stable-Baselines3} were implemented with DQN and PPO for a total of 8,000 timesteps. Table \ref{tab:config} provides an overview of the key configuration parameters for the environment, detailing aspects such as observation shape, action type, and simulation frequency. The evaluation process for the trained agents is outlined in Algorithm 2, which describes the method used to assess the performance of the agents based on metrics such as collisions, successful arrivals, and average speed.


\begin{table}[ht]
\centering
\caption{Environment configuration for intersection-v1.}
\label{tab:config}
\begin{tabularx}{\columnwidth}{X@{\hskip 0.5pt}c}
\toprule
\textbf{Configuration} & \textbf{Value} \\
\midrule
Observation Type             & Grayscale    \\
Observation Shape            & $(128, 64)$  \\
Stack Size                   & 4             \\
Action Type                  & Discrete      \\
Duration (seconds)           & 30            \\
Initial Vehicle Count        & 5             \\
Spawn Probability            & 0.2           \\
Simulation Frequency         & 15            \\
\bottomrule
\end{tabularx}
\end{table}

\begin{algorithm}[H]
\caption{Post-Training Evaluation}
\textbf{Input:} Trained agent (\(\mathcal{A}\)), number of episodes (\(E\))\\
\textbf{Output:} \(\bar{c} = c / E\), \(\bar{s} = s / E\), \(\bar{v} = v / E\)
\setlength{\baselineskip}{12pt}
\begin{algorithmic}[1]
\State Initialize \(c, s, v \)
\For{\(e = 0 : E\)}
    \State \textit{done} \(\gets\) \textbf{False}
    \State \textit{truncated} \(\gets\) \textbf{False}
    \State \(s_0 \gets\) reset the environment
    \State \(a_0 \sim \pi_{\mathcal{A}}(s_0)\)
    \While{not \textit{(done or truncated)}}
        \State \(t \gets t + 1\)
        \State \(s_{t+1}, r_t, \textit{done} \gets\) agent \(\mathcal{A}\) takes an action \(a_t\)

        \State \(v \gets\) append the agent's speed
        \If{\textit{collision}}
            \State \(c \gets c + 1\)
        \EndIf
        \If{\textit{arrived}}
            \State \(s \gets s + 1\)
        \EndIf
    \EndWhile
\EndFor
\end{algorithmic}
\end{algorithm}

\vspace{8cm}
\subsection{Evaluation of Adjusted CLIP Predictions}
\begin{figure}
    \centering
    \includegraphics[width=1\linewidth]{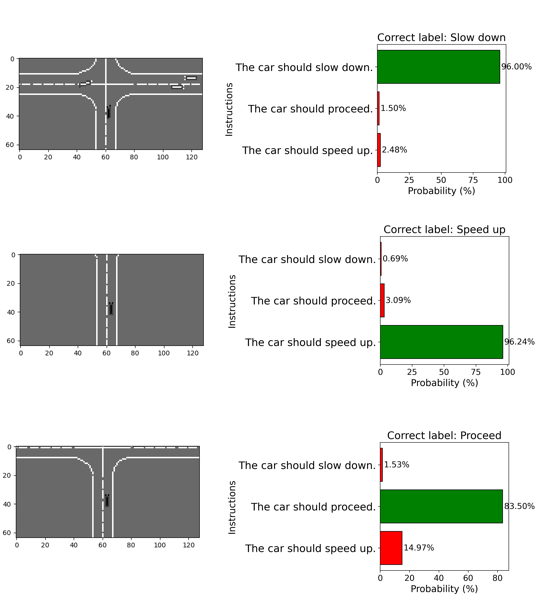}
    \caption{CLIP-based predictions for driving instructions at different intersection scenarios.} Each row corresponds to a distinct intersection situation, with the left panel displaying the intersection scene and the right panel showing the probabilities of three possible actions. The ground truth label is colored green while an incorrect
    prediction is colored red. The probabilities are computed based on the similarity between the intersection scene and the textual instructions using CLIP.
    \label{fig:acc}
\end{figure}

Figure \ref{fig:acc} illustrates the performance of the CLIP-based model in predicting driving instructions for some random intersection scenarios. For each scenario, the green bar represents the probability of the correct label, while the red bars indicate the probabilities of the incorrect labels. The results demonstrate the model's ability to align the intersection scene with the appropriate driving action.

\begin{figure}[H]
    \centering
    \includegraphics[width=1\linewidth]{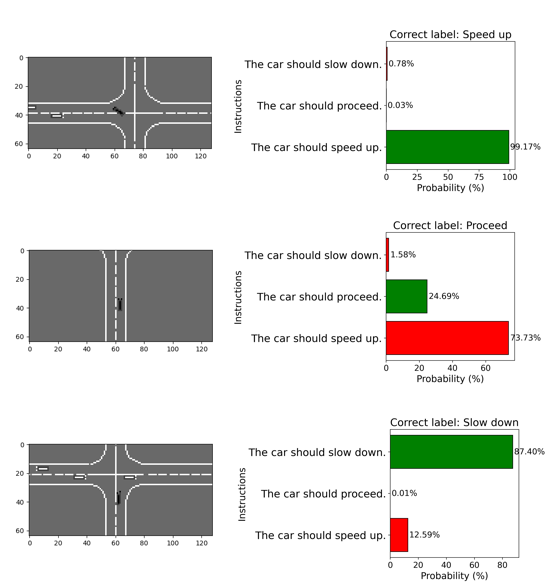}
    \label{fig:enter-label}
\end{figure}

\end{document}